\documentclass[sigconf]{acmart}

\AtBeginDocument{%
  \providecommand\BibTeX{{%
    \normalfont B\kern-0.5em{\scshape i\kern-0.25em b}\kern-0.8em\TeX}}}
    
\usepackage{microtype}
\usepackage{graphicx}
\usepackage{color}
\usepackage{caption}
\usepackage{latexsym}
\usepackage{enumitem}
\setlist{leftmargin=3mm}
\usepackage{amsfonts} 
\usepackage{tikz,amsmath,amsthm,graphicx,subfig}
\usepackage{algorithmic,algorithm,varwidth,booktabs}
\usepackage{array, makecell}
\usepackage{multirow}
\usepackage{tabularx}
\usepackage{mwe}
\usepackage{capt-of}

\newcommand{\model}{\textsc{RLTG}}

\setcopyright{acmcopyright}
\copyrightyear{2020}
\acmYear{2020}
\acmDOI{.}

\acmConference[Woodstock '18]{Woodstock '18: ACM Symposium on Neural
  Gaze Detection}{June 03--05, 2018}{Woodstock, NY}
\acmBooktitle{Woodstock '18: ACM Symposium on Neural Gaze Detection,
  June 03--05, 2018, Woodstock, NY}
\acmPrice{15.00}
\acmISBN{978-1-4503-XXXX-X/18/06}

\begin{document}

\title{Topic-Preserving Synthetic News Generation: \\ An Adversarial Deep Reinforcement Learning Approach}

\author{Ahmadreza Mosallanezhad}
\affiliation{%
  \institution{Arizona State University}
}\email{amosalla@asu.edu}

\author{Kai Shu}
\affiliation{%
  \institution{Illinois Institute of Technology}
}\email{kshu@iit.edu}

\author{Huan Liu}
\affiliation{%
  \institution{Arizona State University}
}\email{huan.liu@asu.edu}

\renewcommand{\shortauthors}{Mosallanezhad et al.}

\begin{abstract}
Nowadays, there exist powerful language models such as OpenAI’s GPT-2 that can generate readable text and can be fine-tuned to generate text for a specific domain. Considering GPT-2, it cannot directly generate synthetic news with respect to a given topic and the output of the language model cannot be explicitly controlled. In this paper, we study the novel problem of topic-preserving synthetic news generation. We propose a novel deep reinforcement learning-based method to control the output of GPT-2\footnote{Our model is based on GPT-2 as GPT-3~\cite{brown2020language} is not publicly available.} with respect to a given news topic. When generating text using GPT-2, by default, the most probable word is selected from the vocabulary. Instead of selecting the best word each time from GPT-2's output, an RL agent tries to select words that optimize the matching of a given topic. In addition, using a fake news detector as an adversary, we investigate generating realistic news using our proposed method. In this paper we consider realistic news as news which cannot be easily detected by a fake news classifier. Experimental results demonstrate the effectiveness of the proposed framework on generating topic-preserving news content than state-of-the-art baselines.

\end{abstract}



\maketitle
\section{Introduction}
Text generation is an important task for Natural Language Processing (NLP). With the rise of deep neural networks such as recurrent neural networks (RNNs) and Long Shot Term Memory (LSTM) cells~\cite{gers1999learning}, there has been a huge performance improvement in language modeling and text generation. 
Text generation has many different applications such as paraphrase generation and data augmentation. 
One important application of text generation in NLP is synthetic news content generation~\cite{zellers2019defending}. 

Recently, social media has proliferated a plethora of disinformation and fake news~\cite{allcott2017social,shu2017fake}. 
Recent advancements on language models such as OpenAI's GPT-2~\cite{radford2019language} allow ones to generate synthetic news based on limited information. For example, models like Generative Adversarial Network (GAN)~\cite{guo2018long} can generate long readable text from noise, and GPT-2~\cite{radford2019language} can write news stories and fiction stories given simple context such as part of a sentence or a topic.
One recent model named Grover~\cite{zellers2019defending} focuses on generating fake news using causal language models, considering different variables such as \textit{domain}, \textit{date}, \textit{authors}, and \textit{headline}. 
While Grover is shown effective, it 
requires many conditional variables to generate relevant news. 
To be able to study the problem of machine generated news on social media, we propose a model to generate realistic synthetic news. This is a crucial task as it enables us to generate synthetic news and study the differences between real and generated synthetic news. As stated before, one major problem in fake news detection systems is that they cannot differentiate between human and machine generated text. Advances in language models enables one to generate fake news and spread it through social media. To tackle this problem, one major step is being able to generate synthetic news. By generating synthetic news, one can study the hidden differences between human and machine generated text, hence preventing disinformation in social media.


Existing methods may fall short when generating realistic news controlled by a specific context. 
In the real-world scenario, fake news usually has a catchy style and should stay on topic to make its audience believe it. For example: ``A shocking new report claims Kourtney Kardashian’s pregnant again''. Thus, it is important to study the problem of topic-preserving and realistic synthetic news generation. Moreover, fine-tuning language models does not help us in this matter as it is non-trivial to enforce topic-preservation on a language model directly. In essence, we address the following challenges: (1) how we can generate news contents similar to human writing; (2) as training a language model is time consuming and needs a lot of resources, how we can use a faster approach to generate news content; and (3) how we can ensure that the generated news content is both realistic and related to a given topic. 

Our solutions to these challenges result in a novel framework
RLTG (\underline{R}einforcement \underline{L}earning-based \underline{T}ext \underline{G}enerator), for generating topic-preserving realistic fake news. 
The proposed framework RLTG consists of three major components: (1) a language model component to generate a probability distribution over a vocabulary for the next word, given a text input; (2) a Reinforcement Learning (RL) component capable of leveraging the language model to control news generation; and (3) a fake news detection module as an adversary to help the RL agent generating realistic fake news contents.
Our contributions are summarized as follows: \\$\bullet$ We study a novel problem of topic-preserving and realistic synthetic news content generation. \\ $\bullet$ We propose a principled framework RLTG which uses language model and deep reinforcement learning along with adversary regularization to generate realistic synthetic news contents. \\$\bullet$ We conduct experiments on real-world datasets using quantitative and qualitative metrics to demonstrate the effectiveness of RLTG for synthetic news generation. 

\section{Related Work}
\label{section_relatedwork}
In this section, we briefly describe the related work on (1) neural news generation; (2) adversarial training; and (3) Reinforcement Learning for Text Generation:
\subsection{Neural news generation}
Text generation is a crucial task in Natural Language Processing and is being used in different applications of NLP~\cite{fan2019strategies, puduppully2019data}. Many early methods for text generation use different techniques to train Generative Adversarial Networks (GAN). As GANs cannot be used for text generation due to the discrete nature of the problem, the early works try to solve the problem of back propagation for updating the generator. Several methods \cite{lin2017adversarial,fedus2018maskgan,guo2018long} have been proposed to alleviate this problem. MaskGAN~\cite{fedus2018maskgan} tries to generate text using both GAN and actor-critic networks. 
Finally, in LeakGAN~\cite{guo2018long}, unlike other GAN-based methods that the discriminator and generator are trained against each other, it uses 
the discriminator to help the generator predict the next word.

Newer methods try to leverage Reinforcement Learning (RL) for the text generation problem. \cite{shi2018toward, fan2018reinforcement} uses inverse RL to solve the problem of mode collapse in GAN, meaning that during the training of GAN, the discriminator becomes too powerful that we cannot train a generator against it. Another work \cite{ke2019araml} models the problem of GAN text generation using RL. They use a reward function as a feedback to the generator. 
While these methods can generate text, they cannot be used for to generate text for a specific domain. Recently, \cite{zellers2019defending} proposes a new causal language model whose goal is to consider different factors such as domain, date, authors, and headline of a news content, to generate fake news. 

To solve the problem of controllable text generation for a specific topic, 
Dathathri et al. proposed to use gradient difference on a pre-trained language model to control its output toward a given attribute. They further use different classifiers to calculate the gradient different~\cite{dathathri2019plug}. Moreover, Hu et al. proposed to use several discriminators in a GAN to control the text generation process~\cite{hu2017toward}. What makes our work different than these methods is that in our work we focus on controlling the output toward a given topic, instead of a given class attribute. 

\subsection{Adversarial training on discrete variables}
Recently there are increasing works focusing on adversarial training over discrete sequence data~\cite{bengio2015scheduled, huszar2015not, yu2017seqgan,lamb2016professor,li2018generative}. Lamb et al.propose providing the discriminator with the intermediate hidden state vectors rather than its sequence outputs, which makes the loss function differentiable for back propagation training~\cite{lamb2016professor}. 

Yu et al. propose a novel method named SeqGAN. They apply a Generative Adversarial Network~\cite{goodfellow2014generative} to discrete sequence generation by directly optimizing the discriminator's rewards using policy gradient reinforcement learning~\cite{yu2017seqgan}. Other approaches use continuous approximation to represent discrete tokens to facilitate the gradient propagation process~\cite{kusner2016gans,hu2017toward}. Continuous approximation uses the Gumbel-softmax function~\cite{jang2016categorical} to transform the one-hot vector 
into a probabilistic vector that is differentiable for training. 

Efforts have been made to generate diverse and high quality text~\cite{zhang2017adversarial, guo2018long}. Guo et al. propose a new method for generating long text using adversarial training. They leverage the hidden states of an adversary as leaked information in order to optimize a GAN to generate long text~\cite{guo2018long}. To broaden the domains of generated text Wang et al. propose a method which uses a multi-class classifier as a discriminator. It further uses multiple generators alongside the discriminator to optimize the model~\cite{wang2018sentigan}. 
Moreover, Zhang et al. propose a novel method, TextGAN, to alleviate the problems of generating text using GAN. They use LSTM as a generator, and a Convolutional Neural Network as a discriminator~\cite{zhang2017adversarial}.

\subsection{Reinforcement learning in text generation}
In the past years, reinforcement learning has shown to be useful in improving model parameters~\cite{li2016deep, li2017paraphrase}. Furthermore, it can be used as a standalone algorithm for different purposes such as dialog or paraphrase generation. Fedus et al. propose a method for overcoming the problems of generating text via GAN. They use reinforcement learning to tune parameters for a LSTM based generator~\cite{fedus2018maskgan}. Zichao Li et al. proposes a method for generating paraphrase using inverse reinforcement learning. They use an RL setting to tune a generator's parameter toward generating paraphrases~\cite{li2017paraphrase}. 
Another inspiring work by Jwei Li et al. shows using reinforcement learning we can build an agent capable of engaging in a two person dialog~\cite{li2016deep}. 
To generate diverse text, Shi et al. propose a method which uses a generator in an RL setting. The difference between their work and other similar works is that they also change the parameters of the reward function during the training process~\cite{shi2018toward}. 

Inspiring by these methods, we study the problem of synthetic news generation using RL. While these methods use RL to update a model's parameters, in our work, we focus on using RL alongside a language model to control its output towards news generation. In this novel model, we train an agent to use a language model's output to control the generation process, thus generating realistic synthetic news.
\section{Problem Statement}
Let $\mathcal{X} = \{\mathbf{(S_0^1, x_1), (S_0^2, x_2), ..., }$
$\mathbf{(S_0^N, x_N)}\}$ denote a set of $N$ news with topic $\mathbf{S_0}$ and content $\mathbf{x}$. Both topic $\mathbf{S_0} = \{w_0, w_1, ..., w_k\}$ and news content $\mathbf{x} = \{w_0, w_1, ..., w_l\}$ consist of several words $w$. We consider topic as the news title or the first few words of a news content. In general, $S_t$ shows the generated text at time $t$. 

In this paper our goal is to generate synthetic news content $\mathbf{S_T}$ given topic $\mathbf{S_0}$ 
The generated news content $\mathbf{S_T}$ should be related to the given topic $\mathbf{S_0}$ and it should be realistic. Here we define realistic synthetic news as a news content that cannot be easily detected as fake news using a classifier. 
Here, we study the following problem:

\textbf{Problem 3.1.} Given a set of news dataset $\mathcal{X}$, learn a reinforcement learning agent $F$ that can generate news content $\mathbf{S_T}$ based on a given topic $\mathbf{S_0}$ such that: (1) $\mathbf{S_T}$ is related to the given topic $\mathbf{S_0}$; and (2) $\mathbf{S_T}$ is realistic and cannot be easily detected as fake.

\section{Proposed Model - {\model}}
\label{section_model}
In this section, we discuss the adversarial reinforcement learning-based synthetic news content generator. The input of this model is topic $\mathbf{S_0} = \{ w_1, w_2, ..., w_k \}$. Our model, then, generates a new sequence $\mathbf{S_T} = \{w_1, ..., w_k, w_{k+1}, ..., w_T\}$ which is the final generated news content. Our model consists of several components:
(1) a language model component which is in charge of generating a probability distribution over vocabulary words
; (2) an RL component which will select a word based on the language model's output; and (3) an adversarial component which will help the RL agent choosing proper words from the language model's output.
First, we go through news content generation using adversarial RL, then, we discuss using adversary to generate realistic fake news.

\subsection{Reinforced News Generation}
Existing language models are proposed to generate general or domain-specific texts~\cite{devlin2018bert,radford2019language}. Although we can fine-tune these models (i.e. fine-tuning GPT-2) to our need using a dataset, we do not have control over its output as we cannot enforce topic-preservation or realistic synthetic news generation on the model. 
Following the success of Reinforcement Learning (RL)~\cite{shi2018toward}, we propose an adversarial RL method to control the generated output of a language model. In recent works, RL has been used to update a model's parameters~\cite{li2016deep, li2017paraphrase}. In this work we explore a new direction by using RL as a standalone component in order to leverage the language model's output to generate text. The main advantage of using RL in this way is that we can use non-differential metrics in the reward function in order to generate a more readable text. Moreover, it enables us to have more control on the output of the language model by leveraging adversaries or changing the reward function.

In adversarial RL an agent keeps interacting with a defined environment to learn an optimized action selection policy $\pi(s)$ for each state. An RL agent is trained to choose the next word $w$ for current generated news $\mathbf{S_t}$ according to a reward function and an adversary.

\begin{figure}[t!] \vspace{-10pt}
    \centering
    \includegraphics[scale=0.13]{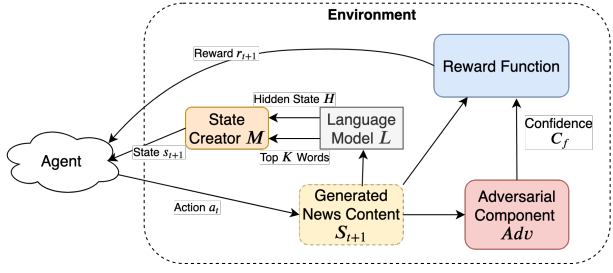}
    \caption{The architecture of our proposed model. The environment consists of a language model $L$, reward function, state creator $M$, and an adversary $Adv$. The agent keeps interacting with the environment to find the best action $a$ for each given state $s$.}
    \label{fig:model}
\end{figure}

Figure~\ref{fig:model} shows the high-level structure of RLTG. In this model, the adversarial reinforcement learning agent gets an state $s_t$ as input, then returns action $a_t$ which indicates an index to one of the top words from the language model $L$'s output. Each interaction between the agent and the environment creates an experience tuple $(s_t, a_t, s_{t+1}, r_{t+1})$, meaning that the agent chose action $a_t$ in given the state $s_t$. After action $a_t$, the state will change to $s_{t+1}$ and the environment returns reward $r_{t+1}$. This tuple is then used to train the agent. An RL model relies on four main parts: environment, state, action, and reward function:

$\bullet$ \textbf{\textit{Environment}} is where the RL agent interacts with to learn the best action for each state. In our problem, the environment includes a language model $L$, an adversary \textsc{Adv}, and a state creator component $M$. The language model $L$ takes an input text and returns a probability distribution over vocabulary $P \in {\rm I\!R}^{1 \times |V|}$ and hidden states $H \in {\rm I\!R}^{1 \times e}$, where $e$ indicates the embedding size. The adversary \textsc{Adv} gets an input text and returns a score for the reward function. Finally, the state creator $M$ gets the outputs of the language model as input and returns a vector $\mathbf{s} \in {\rm I\!R}^{1 \times |s|}$ ($|s|$ shows state size) which acts as the input state $s$ for the agent. 

$\bullet$ \textbf{\textit{State}} shows the agent's current situation. The agent uses the state to determine 
a subsequent optimal action. The state is the output of the state creator component $M$. 
As our goal is to select best next word for the current generated news $\mathbf{S_t}$ at time $t$, the state should contain information about both the context of the current generated news $\mathbf{S_t}$, and information about the next word choices. To this end, we design two separate neural networks $AE^1$ and $AE^2$ to encode these information. $AE^1$ is used to create the context vector $c_g$ using hidden state $\mathbf{H}$ from the language model $L$'s output, while the $AE^2$ is used to create a context vector $c_w$ given previous top $K$ words of the language model $L$'s output. For both cases, we train and use autoencoders~\cite{ap2014autoencoder}. Autoencoder is an unsupervised neural network which learns how to compress and encode data, and then how to reconstruct the input using the encoded data. An autoencoder has two components, an encoder and a decoder. The encoder takes an input and returns a vector $\mathbf{v}$ which is interpreted as the context vector, containing important information about the input. The decoder is the reverse of the encoder and having the encoded vector $\mathbf{v}$, it tries to reconstruct the original input to the network. After training an autoencoder, we can use the context vector $v$ containing important information about the input~\cite{ap2014autoencoder}.

In our method, the first autoencoder $AE^1$, gets the hidden state $\mathbf{H}$ as input and returns the reconstructed hidden state $\mathbf{H'}$. This autoencoder uses Multi Layer Perceptron (MLP) networks as both encoder and decoder. Our goal in this autoencoder is to reduce dimension of the hidden state $\mathbf{H}$. After training this autoencoder on a set of hidden states $\mathbf{H}$, we get the output of the encoder as context vector $\mathbf{c_g}$.

The second autoencoder $AE^2$, inspired by \cite{conneau2016very}, uses Convolutional Neural Network (CNN) as both encoder and decoder. To this end, each word from top $K$ words is passed through an embedding layer to convert it to a vector $\mathbf{w} \in {\rm I\!R}^{1 \times e}$. The embedded words $\mathbf{w}$ are then concatenated to form a matrix $m$ with size of $(K \times e)$. After training this autoencoder using different top $K$ words, we consider the output of the encoder as the context vector $\mathbf{c_w}$. 
Having both context vectors $c_g$ and $c_w$, we then concatenate both context vectors $\mathbf{s} = Concat(\mathbf{c_g}, \mathbf{c_w})$ to create the state for the RL environment.

$\bullet$ \textbf{\textit{Actions}} indicate the agent's response to a given state $s$. As the agent's goal is to select words, the action set $A$ can be equal to choosing a word from the vocabulary set $\mathcal{V}$. By choosing $\mathcal{V}$ as the agent's action set, we encounter two problems: First of all, it takes a long time to train an agent on a large action set as the agent should try every action in order to find the best action $a$ for each given state $s$ \cite{dulac2015deep}. Secondly, by having a large action set $A$, the agent may not be able to see every state-action set $(s, a)$ in a limited time, hence, it may result in underfitting~\cite{dulac2015deep}. To solve these problems, we make use of the language model $L$'s output. One of the outputs of the language model is the probability distribution over vocabulary $\mathcal{V}$. The probability distribution indicates what are the likely best options to sample the next word for a given text $\mathbf{S_t}$. In this paper, we select top $K$ words of the probability distribution as the action set, leading to a small action set. 

$\bullet$ \textbf{\textit{Reward Function}} evaluates agent's actions for each given set $(s_t, a_t)$. During training, the agent uses the reward function to learn the best strategy for selecting actions. In this paper, the goal is to generate a synthetic news content which is related to a given topic. To this end, we use cosine similarity to measure similarity between the given embedded topic $\mathbf{S_0}$ and the current generated synthetic news $\mathbf{S_t}$. The reason behind using the embedded topic and generated synthetic news at time $t$, is that using the exact words in the Cosine similarity function may result in an agent which chooses topic word to maximize this similarity:
\begin{equation}
    CosineSim(\mathbf{S'_0}, \mathbf{S'_t}) = \frac{\mathbf{S'_0} \cdot \mathbf{S'_t}}{ ||\mathbf{S'_0}|| \cdot ||\mathbf{S'_t}||}
    \label{eq:cosinesim}
\end{equation}
where $S'$ is the embedded topic/news using the language model $L$. We use the language model $L$'s hidden state $H$ as the embedding for an input text as it shows the context of an input text~\cite{budzianowski2019hello}.

Furthermore, for generating news content, the model should pay attention to the writing style of news content. In this paper, we consider style as having a similar word sequence as the reference news. To this end, for a given synthetic generated news $\mathbf{S_t}$, we calculate the BLEU score~\cite{papineni2002bleu} between $\mathbf{S_t}$ and news contents $\mathcal{X}$ to maintain news style. The BLEU score simply measures how many words overlap between the generated news $\mathbf{S_t}$ and the reference news contents $\mathcal{X}$. As BLEU metric gives higher scores to similar sequential words, it can be used as a fluency metric in the designed reward function. The reward function is as follow:
    \begin{equation}
        r_t = \alpha CosineSim(\mathbf{S^{'i}_{0}}, \mathbf{S'_t}) + \beta BLEU(\mathbf{S_t}, \mathcal{X})
    \end{equation}
    where $\alpha$ controls the contribution of Cosine similarity term, and $\beta$ controls the contribution of BLEU score. 
    

\begin{figure}[t!] \vspace{-15pt}
    \centering
    \includegraphics[scale=0.14]{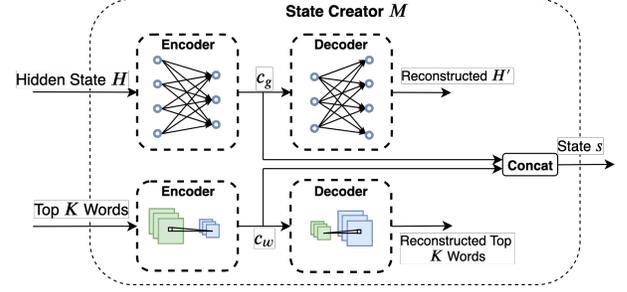}
    \caption{The architecture of state creator $M$. It leverages two autoencoders to create state $s$ for the RL agent.}
    \label{fig:state}
\end{figure}

\subsection{Using Adversary to Control Synthetic News Generation}
Till now, we can generate a text related to a given topic. 
To ensure that the generated news is realistic enough, we use a fake news detection component as an adversary to determine whether the generated news is considered fake or not. Thus, we add an additional term to the reward function:
\begin{align}
    \label{eq:reward_2}
    r_t =& \alpha CosineSim(\mathbf{S^{'}_0}, \mathbf{S'_t}) + \beta BLEU(\mathbf{S_t}, \mathcal{X}) + \lambda (1 - C_{f}(\mathbf{S_t}))
\end{align}
    
Where $C_f \in [0, 1]$ is the confidence of the fake news classifier given an input, and $\lambda$ shows the importance of this term. The confidence shows the probability of a news content being fake. 

For training the agent, we use news dataset $\mathcal{X}=\{\mathbf{(S_0^1, x_1)}, \mathbf{(S_0^2, x_2)}$ 
$, ..., \mathbf{(S_0^N,x_N)}\}$ in which $\mathbf{S_0^i}$ shows the topic of $i^{th}$ news and $\mathbf{x_i}$ shows the content of that news. During training, the agent chooses an action $a_t$ leading to selecting word $w_{t} \in \mathcal{V}$, which is then added to current generated news $\mathbf{S_{t}} = \{ w_1, w_2, ..., w_k, ..., w_t \}$ to generate $\mathbf{S_{t+1}} = \{w_1, ..., w_k, ..., w_t, w_{t+1}\}$. The modified text $\mathbf{S_{t+1}}$ is then passed to the adversary $C_f$ and the reward function to calculate the reward value $r_{t+1}$ considering news content $\mathbf{x}$. Furthermore, the modified text $\mathbf{S_{t+1}}$ is passed to the language model $L$. Using the outputs of the language model $L$ the environment generates next state $s_{t+1}$. In the following we discuss the details of using adversarial reinforcement learning.

In adversarial reinforcement learning, the goal is to learn an action policy $\pi(s)$ which leads to maximum amount of accumulated reward $R = \sum_{t=0}^{t=T} r_t$ where $T$ is the terminal time. To find the best action selection policy $\pi(s)$, we use experiences in form of $(s_t, a_t, s_{t+1}, r_{t+1})$ to train the agent. There are different algorithms to train an agent. Policy gradient and Q-Learning are two popular algorithms for training an agent~\cite{sutton1998introduction}. In this paper we use Deep Q-Learning which is an advanced variant of Q-Learning. 

In Deep Q-Learning (DQL), the agent uses a neural network as a function approximator to find action it should select regarding to a given state $s$. The input of this neural network is state $s$ and the outputs are the values for $(s, a_i)_{i=0}^{|A|}$ where $|A|$ is the number of actions. In DQL, the goal is to learn the following function:
\begin{equation}
    Q^*(s_t, a_t) = {\rm I\!E}_{s_{t+1}} [ r_{t+1} + \gamma max_{a'} Q^* (s_{t+1}, a') ]
    \label{eq:dql_optim}
\end{equation}
where Q-function $Q(s, a)$ returns the expected accumulated reward $R$ if the agent selects action $a$ in response to state $s$ and $Q^*(s, a)$ denotes the optimal Q-function which returns the maximum possible accumulated reward $R$ using the optimal policy $\pi(s)$. In this formula, the future rewards are discounted using the $\gamma$ parameter. We adjust $\gamma$ with respect to the importance of future rewards. 

In practice, it is not feasible to estimate $Q^*(s, a)$ in Equation~\ref{eq:dql_optim}. To overcome this problem, we use a function approximator to estimate the Q-function $Q^*(s, a) 	\cong Q(s, a;\theta)$. As neural networks are excellent function approximators~\cite{cybenko1989approximation}, DQL leverages a neural network with parameters $\theta$ called Deep Q-Network (DQN) to find the Q-function $Q(s, a;\theta)$ by minimizing the following loss function:
\begin{equation}
    L(\theta) = {\rm I\!E}_{s_t, a_t, s_{t+1}, r_{t+1}} [(y - Q(s, a;\theta))^2]
    \label{eq:dqn_loss}
\end{equation}
where $y$ is the target Q-value calculated using Equation~\ref{eq:y}:
\begin{equation}
    y = {\rm I\!E}_{s_{t+1}} [r_{t+1} + \gamma Q(s_{t+1}, a';\theta')]
    \label{eq:y}
\end{equation}
where $\theta'$ is the DQN's parameters from the previous iteration.

Finally, we update the DQN parameters using the derivation of Equation~\ref{eq:dqn_loss} with respect to $\theta$:
\begin{align}
    \centering
    \label{eq:update-dqn}
    \nabla_{\theta} L(\theta) =& {\rm I\!E}_{s_t, a_t, s_{t+1}, r_{t+1}} [(r + \gamma max_{a'} Q(s_{t+1}, a';\theta') - \\\nonumber
    &- Q(s_t, a_t;\theta)) \nabla_{\theta} Q(s_t, a_t;\theta)]
\end{align}

In this paper, we have specifically used DQL with memory replay and two networks as target and policy respectively. The memory replay helps the agent to remember past experiences. The training algorithm is presented in Algorithm~\autoref{alg:alg1}\footnote{The dataset and source code will become available upon acceptance.}.

\begin{algorithm}[t]
	\caption{The Learning Process of \textsc{RLTG}}\label{alg:alg1}
	\begin{algorithmic}[1]
		\REQUIRE~~ $L$, $\epsilon$, $T$, $M$.
		\STATE Initialize replay memory $R$, environment, policy, and target networks
        \WHILE{training is not terminal}
            \STATE $H, top\-K \gets L(topic)$
            \STATE $s_t \gets M(H, top\-K)$
            \FOR{$t \in \{0, 1, ..., T\}$}
                \STATE Choose action $a_t$ using $\epsilon$-greedy 
                \STATE Perform $a_t$ on $s_t$ and get $(s_{t+1}, r_{t+1})$
                \STATE $R \gets R + (s_t, a_t, r_{t+1}, s_{t+1})$
                \STATE $s_{t} \gets s_{t+1}$
                \FOR{$(s, a, s', r) \in$ sampled mini-batch $ b$ from $R$}
                    \STATE Update DQN weights using Eq.~\ref{eq:update-dqn} w.r.t. policy and target networks
                \ENDFOR
                \IF{exchange condition met}
                    \STATE Exchange weights between policy and target network
                \ENDIF
            \ENDFOR
        \ENDWHILE
		
	\end{algorithmic}
\end{algorithm}
\section{Experiments}
\label{section_experiments}

In this section, we conduct experiments to evaluate the performance of our method. In these experiments, we try to answer the following questions: \textbf{Q1:} How well our method can generate news content regarding to a given topic? \textbf{Q2:} How fluent is the generated synthetic news using RLTG? and \textbf{Q3:} How well RLTG generates synthetic news content in comparison to other existing methods.

To answer the first question, we consider Cosine similarity, while for Q2, we use ROUGE-L metric. Finally, to answer the last question Q3, we compare generated news from RLTG to other existing text generation methods and perform human evaluation.

\subsection{Data}
We utilize FakeNewsNet dataset~\cite{shu2018fakenewsnet} to fine-tune GPT-2 and train our model. This dataset consists of news data $\mathcal{X}$ from two different platforms \textit{GossipCop} and \textit{Politifact}. GossipCop is a fact-checking website which reports on celebrity news. Politifact is a similar platform which checks the truth of political news and reports. In this dataset, news are classified into \textit{real} or \textit{fake}. \textit{Politifact} contains $2,645$ true and $2,770$ fake news, while the \textit{GossipCop} includes $3,586$ true and $2,230$ fake news respectively.
In this paper, we consider the first few words of each news $x_i$ content as topic $\mathbf{S_0}$. 

\begin{table*}[t!]
\centering
\begin{tabularx}{\linewidth}{ c X }
    \toprule
      Model & Text \\\hline\\ [-1.8ex]
      
      RLTG & \begin{minipage}{0.85\textwidth}\textbf{the wedding of prince harry and meghan markle} was held with royal family guests attending the ceremony. After nearly four hours on site, they met again with wedding guests from a host community to gather for one particular special day. In May there would been even-clothing on all royal guests arriving as many couples wore the customised royal gown...\end{minipage} \\\hline \\ [-1.8ex]

      Fine GPT-2 & \begin{minipage}{0.85\textwidth}\textbf{the wedding of prince harry and meghan markle} was held on the grounds of St. George's Chapel in Windsor, England, on May 19, 1814. The bride and groom were Prince William and Kate Middleton, and the groom was Prince Charles. The bride...\end{minipage}\\\hline \\ [-1.8ex]
      
      GPT-2 & \begin{minipage}{0.85\textwidth}\textbf{the wedding of prince harry and meghan markle} was held on the wedding of prince harry and meghan markle was held on the day of the wedding of prince harry and meghan markle was held on the day...\end{minipage} \\ \hline\\ [-1.8ex]
      
      Grover & \begin{minipage}{0.85\textwidth}\textbf{the wedding of prince harry and meghan markle}.  Like many of the gala trends we celebrate today, the wedding of prince harry and meghan markle has taken center stage at the MWC. The mom and pop couples still roping in floral designs, staging weddings, and of course, speaking on stage at the hands of this mortal pair.Seeing their horses pushing their new princesses towards them on display in an exhibition hall makes for a highlight, especially with a gargantuan fairytale looming on the horizon. Julianne Ros was the one to collect a whole circus of gowns, each a special cover up for a guy and a woman. \end{minipage} \\ \hline \hline \\ [-1.8ex]
      
      RLTG & \begin{minipage}{0.85\textwidth}\textbf{kim kardashian and kanye west are reportedly getting married in florence}, Italy! Kim confirmed it this evening after receiving numerous texts during Sunday. We will certainly keep up this morning's news, as long as this story reaches our lips...\end{minipage}\\\hline\\ [-1.8ex]
      
      Fine GPT-2 & \begin{minipage}{0.85\textwidth}\textbf{kim kardashian and kanye west are reportedly getting married in florence}. The couple, who have been dating since 2015, have been spotted together in the city of L.A. and are reportedly spending time together. The couple, who have been dating since 2015, have been spotted together in the city of L.A. and are reportedly spending time together...\end{minipage}\\ \hline\\ [-1.8ex]
      
      GPT-2 & \begin{minipage}{0.85\textwidth}\textbf{kim kardashian and kanye west are reportedly getting married in florence}. The couple is reportedly planning to have a baby together. The couple is reportedly planning to have a baby...\end{minipage}\\ \hline\\ [-1.8ex]
      
      Grover & \begin{minipage}{0.85\textwidth}\textbf{kim kardashian and kanye west are reportedly getting married in florence}. According to BeBe, whose team is developing ready-made phones. Kim Kardashian West and Kanye West have called it quits on their engagement. "Kim Kardashian and Kanye West announce divorce at 38 after worst divorce in history!" That's according to the site, which also notes that the pair met at a New York karaoke contest and that they intend to have a daughter together.\end{minipage}\\ \bottomrule
\end{tabularx}
\caption{\label{table:generated} Sample generated news given a news topic. The topic is in \textbf{bold} text.}
\end{table*}

\subsection{Implementation Details}
In this part we go through the parameters and implementation details of RLTG.
In our model, we use a fine-tuned GPT-2 language model as $L$. To fine-tune the GPT-2 language model, we first load a pre-trained "GPT-2 medium", then we use FakeNewsNet dataset for $5$ iterations to fine-tune the language model. Note that this language model has $12$ hidden layers. Each hidden layer returns a tensor with size of (batch size, sequence length, hidden size), where the hidden size in "GPT-2 medium" is $768$. 

As it is mentioned in the proposed method, the RL agent has a neural network which acts as a function approximator. This network gets a state as input and returns the Q-value for each $(s, a_i)_{i=1}^{K}$ set. This network has $3$ layers. The first hidden layer has $1024$ nodes, the second and third layer has $512$ and $256$ nodes respectively. The output size of this network is equal to the number of actions. In this paper, the number of actions is $50$, meaning that the agent chooses between the top $50$ words of GPT-2's output probability. The reason we chose $50$ is that among values $\{10, 25, 50, 75\}$, it showed a better reward performance among others, with $K = 75$ having similar performance. Moreover, the the output size of the DQN network is equal to the size of state $\textbf{s}$. To construct state $\textbf{s}$, as in Figure~\ref{fig:state}, we have trained 2 autoencoders and concatenate the output of the each encoder to create the state. The first autoencoder is considered for extracting the context of generated news using hidden state $H$. 
This autoencoder uses Multi-Layer Perceptron (MLP) to encode and reconstruct the hidden state $H$. The output of encoder part has $256$ nodes. The second autoencoder uses Convolutional Neural Networks (CNN) to extract information about best words positions. The encoder of this autoencoder has an output layer with size of $128$. The final size of state $\mathbf{s}$ is $384$. 

\begin{table}[ht!]\vspace{-10pt}
    \centering
    \small
    \begin{tabular}{lcccccc}
    \toprule
      & {RLTG} & GPT-2 & FTGPT-2  & Grover & SeqG & RL\\\midrule
    Similarity & \textbf{0.342} & 0.176 & 0.241 & 0.313 & 0.301 & 0.153\\
    Perplexity & \textbf{14.8} & 22.3 & 19.8 & 15.3 & 17.4 & 30.4\\
    ROUGE-L & \textbf{28.4\%} & 23.1\% & 24.6\% & 27.3\% & 21.5\% & 17.2\%\\
    \bottomrule
    \end{tabular}
    \caption{\label{table:perplex} Topic similarity ($\uparrow$ better), perplexity ($\downarrow$ better), and ROUGE-L score ($\uparrow$ better) based on model's generated news.}
\end{table}

We train the RL agent on randomly selected topics from our dataset. The agent can choose between top $K=50$ words from $L$'s output. We train the RL agent for $50000$ episode. Each episode has a terminal time of $T = 50$. As the final generated news is important for us, we select a high discount factor $\gamma = 0.9$. As it is mentioned in the proposed method section, we use Deep Q-Learning to train our RL agent. In this algorithm we construct a memory with size $10000$ to save the experiences $(s_t, a_t, s_{t+1}, r_{t+1})$. Each experience means that the RL agent chose action $a_t$ in state $s_t$. The selected action $a_t$ resulted in transition to a new state $s_{t+1}$ and the environment returned a reward $r_{t+1}$. We then use the memory array to update our model using Equation~\ref{eq:dqn_loss}. The batch size for sampling experiences from memory is $32$. During the action selection during training, we use $\epsilon$-greedy to choose action $a_t$. This algorithm considers a random action with probability of $\epsilon$ and chooses the best action based on Q-values with a probability of $1 - \epsilon$. We use the following decay function to lower the value of $\epsilon$. This function lower the $\epsilon$ according to the number of past iterations and exponentially decreases it by a constant rate $\epsilon = \epsilon_{min} + (\epsilon_{max} - \epsilon_{min}) e^{ \frac{-steps}{decay\_rate}}$ where $steps$ is the number of past iterations and $decay\_rate$ controls how fast the $\epsilon$ should decrease. We use $\epsilon_{max} = 0.98$, $\epsilon_{min} = 0.02$ and the decay rate equal to $5,000$.
 As for the reward function parameters, we set $\alpha=\beta=\lambda=0.5$. In this case $r \in [0, 1.5]$. 

\begin{figure} \vspace{-20pt}
    \centering
    \includegraphics[scale=0.18]{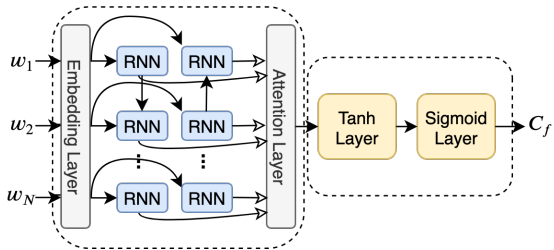}
    \caption{The architecture of fake news classifier. It uses bi-directional GRU and an attention layer to create context vector, then uses a three layer neural network to classify the news as fake or true.}
    \label{fig:classifier}
\end{figure}

As illustrated in Figure~\ref{fig:model}, we use a fake news classifier as an adversary to calculate the value of reward function. The architecture of the fake news classifier is shown in Figure~\ref{fig:classifier}. The hidden size of bi-directional GRU is $128$, resulting in a context vector of $256$. The neural network classifier has an input size of $256$, hidden size of $128$, and output size of $1$. We train this classifier before training the agent using Binary Cross Entropy (BCE) loss function. As DQL has a variance during training, we train our model $5$ times independently, then we select the agent with the highest average rewards. 

\begin{table*}[t]\vspace{-10pt}
\centering
\begin{tabularx}{\linewidth}{ X | c}
\toprule
  Generated news (The \textbf{bold} part is the given topic) & Mean score \\\hline\\ [-1.8ex]
  
  \textbf{The star who was accused of rape late last year} is getting on an international show: ``because I will bring awareness by supporting our families from India for this year on and up! I need every bit it might. this week has turned one world back". Pranta had previously made some serious noises over India-friendly topics ... & 2.33 \\\hline\\ [-1.8ex]

  
  
  
  \textbf{share fans of netflix’s cult favorite scifi drama series sense8} have a very special gift of nostalgia from them that can never truly forget. a beloved franchise in some small but undeniable shape. So this, we know how fans would look after that they would not enjoy being seen. it will become apparent soon when people who enjoy being on such shows, may also like to share these experiences on facebook to have them see those great and unique stories. & 2.25 \\\hline\\ [-1.8ex]
  
  \textbf{jade is my first friend to ever be pregnant in} her mid 30ies. I can feel good about the baby! she will make an excellent wife. I just don't expect a kind person or family who wants the support of an anesthesia for that to occur to a man with diabetes. & 2.00\\\hline\\ [-1.8ex]
  
  
    \textbf{imdbcom inc takes no responsibility for the content or accuracy} of its claims. It has taken place today (August 21), when an article appeared at Wired, which says: "there will be one man with guns on their faces who at a certain level who can take on ISIS". It has gone into further information on what constitutes terrorist threats (or just about all terrorist activity). & 2.50 \\\hline\\ [-1.8ex]

    \textbf{robert pattinson says he was just kidding around about being} asked by anorexist about a possible relationship, "so we were like, I don't know if he's a guy that I want. We're just trying something out, and we just don't have that. So he said I don't want that. He was like a little boy and I don't want it." I have no plans that are to go into this story. & 2.67 \\

  
\bottomrule
\end{tabularx}
\footnotesize
\caption{\label{table:examples} Sample generated news and their average human evaluation scores given a news topic using RLTG.}
\end{table*}

\subsection{Experimental Design} 
We use different baselines for comparison. As our proposed model is based on the OpenAI's GPT-2 language model, we use this language model alone as a baseline to determine how using an RL agent on this model can improve its results. Furthermore, we also include the RL agent alone as a baseline to generate synthetic news. As our main goal, generating synthetic news content, is close to the Grover~\cite{zellers2019defending}, we have selected this work as a baseline as well. Finally we select the SeqGAN method as it incorporates GAN with reinforcement learning. Following is the description of the baselines:
\begin{itemize}[leftmargin=*]\itemsep0em
    \item \textbf{GPT-2 \cite{radford2019language}:} a language model capable of generating long text. This language model is based on transformers and has three different variations based on it's number of layers and parameters: small (117M parameters) , medium (345M parameters), and large (774M parameters). In this paper we use GPT-2 medium as fine-tuning it needs less resources.  
    
    \item \textbf{Fine-tuned GPT-2 (FTGPT-2):} the same as GPT-2, but it has been fine-tuned using FakeNewsNet dataset.  
    
    \item \textbf{RL:} in this baseline we use RL technique without using a language model to train an agent. In this case, all components except the \textit{actions} are the same as the proposed RLTG method. The action set in this baseline is all word in the vocabulary set $V$.  The training process of this baseline is similar to our model.
    
    \item \textbf{Grover \cite{zellers2019defending}:} a conditional language model which can generate text based on given parameters: domain, date, authors, and headline. The goal of Grover is to generate news content based on different parameters. While the results are promising, it seems this language model is very dependent on \textit{domain} parameter which we will explore during our evaluation.
    
    \item \textbf{SeqGAN (SeqG) \cite{yu2017seqgan}:} is a text generation method which models data generator as a stochastic policy in reinforcement learning. They then use policy gradient method to train their model.
    
\end{itemize}

\subsection{Experimental Results}
In this subsection, we evaluate our model's performance: 

\textbf{Topic Similarity (Q1).} To answer the first question (\textbf{Q1}), we use cosine similarity as in Equation~\ref{eq:cosinesim} to calculate the similarity between embedding of the given topic $\mathbf{S_0}$ and the generated news $\mathbf{S_T}$. The reason we use embedding for calculating the similarity is that we do not want our RL agent to exactly select topic words to maximize its reward. In this way, the agent tries to choose words in order to maximize the context similarity between both topic and the generated text. As our model can generate text for different topics, it is not feasible to train a topic classifier for this evaluation. For fair comparison, we use a fixed sentence length of 200. 

\begin{figure*}[ht!] \vspace{-25pt}
    \centering
    \subfloat[\textbf{The confidence of fake news classifier from \\ different episodes during training process.}]{\includegraphics[width=.33\linewidth]{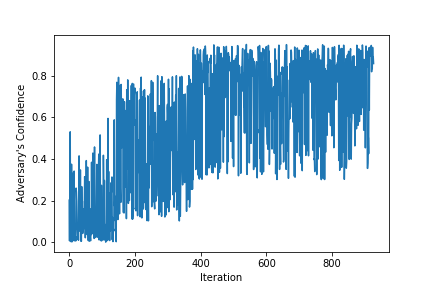}}
    \subfloat[\textbf{Average of agent's reward from episodes \\ $1$ to $4000$.}]{\includegraphics[width=.30\linewidth]{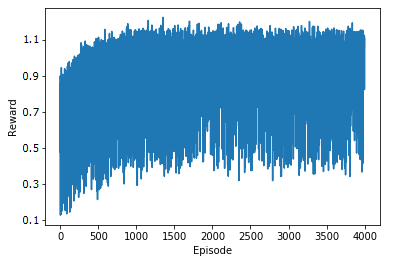}} \\
    \caption{Impact of different reward function parameters on RLTG's ROUGE-L score.}
     \label{fig:rewards-conf}
\end{figure*}

\begin{figure*}[ht!] \vspace{-20pt}
    \centering
    \subfloat[\textbf{Impact of $\alpha$}]{\includegraphics[width=.33\linewidth]{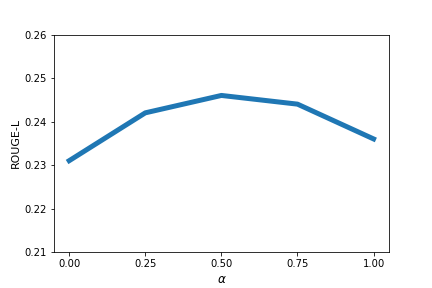}}
    \subfloat[\textbf{Impact of $\beta$}]{\includegraphics[width=.33\linewidth]{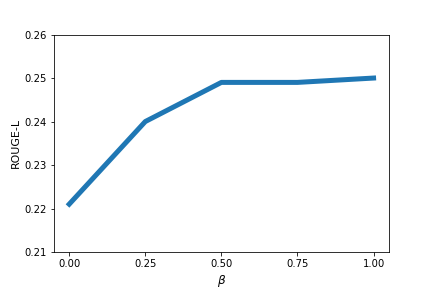}}
    \subfloat[\textbf{Impact of $\lambda$}]{\includegraphics[width=.33\linewidth]{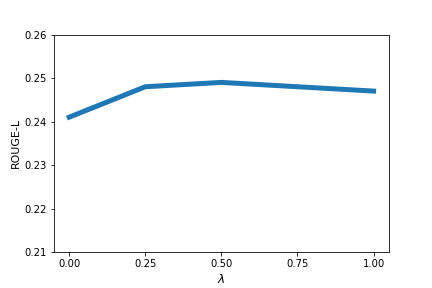}} \\
    \caption{The adversary's confidence and RL agent's rewards show the learning process.}
    \label{fig:params}
\end{figure*}

Table \ref{table:perplex} shows the performance of RLTG against other baselines. As our model considers topic similarity in training, it can outperform other baselines. Although fine-tuned GPT-2 falls behind RLTG and Grover, it has achieved a high similarity comparing to other methods. The reason behind this is that it tends to repeat itself. Note that the performance of the RL baseline is behind all models. The reason behind it is that the action set in this case is very large and the agent cannot converge easily. Furthermore, it shows that using a language model to narrow down the possible actions can have a huge impact on training the model.

\textbf{Fluency Test (Q2)}. To answer this question we use both perplexity and ROUGE-L metrics. Perplexity may not be suitable for showing the effectiveness of a model in open-domain text generation~\cite{liu2016not}, but in our case, we focus on news generation limited in domain. Table~\ref{table:perplex} also shows the results for fluency test. Lower perplexity means the generated news is more concentrated and it is less variant. 
Furthermore, the ROUGE-L score applies Longest Common Subsequence between the news contents $\mathcal{X}$ and generated news content $\mathbf{S_T}$ to calculate the final score. We chose this metric as using BLEU score for evaluation is not fair since our method is trained to give a high BLEU value. ROUGE-L simply measures how many words from the reference sentences have appeared in the generated news. As it gives higher scores to sequential words, it can be used as a fluency metric. In this paper we use the FakeNewsNet dataset as the reference sentences. Higher ROUGE-L score means the generated news is more fluent. 

\begin{figure} \vspace{-10pt}
    \centering
    \includegraphics[width=.8 \linewidth]{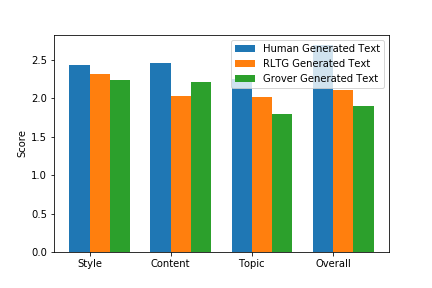}
    \caption{Human evaluation results. Each participants evaluated each articles based on its style, topic similarity, content quality, and overall evaluation. Higher score is better.}
    \label{fig:human_eval}
\end{figure}

\textbf{Generated news (Q3)}. For this part, we generate text for different given topics and compare the results with other baselines. As it is shown in Table~\ref{table:generated}, both base and fine-tuned GPT-2 tends to repeat themselves after generating several words. This problem can be bypassed by resetting the hidden variables. Grover can generate readable and related text. But when it comes to different domains, the generated news can change dramatically. In this situation, the best case is when the domain is related to the given topic. Furthermore, as the goal is to generate synthetic news content, the output of Grover does not seem to be related to news. In this test we excluded the results of the RL baseline, because the generated text was repeated and not readable. 

To further investigate the quality of generated text, we conducted a human study to answer four main question about both human generated, and machine generated text. In this human study, we asked the participants to give a score from 1 to 3 about topic similarity, writing style, content quality, and overall evaluation of the given text. The designed questionnaire is available in the \autoref{appx:A}. We considered best performing models, RLTG and Grover. We included 75 articles, 25 human generated, 25 RLTG generated, and 25 Grover generated. The results are provided in Figure~\ref{fig:human_eval}. Considering the presented results and the generated text from Table~\ref{table:examples} we conclude that RLTG is capable of generating realistic news.

In this part we study the effect of using a fake news detection classifier as an adversary to see if the generated news is realistic enough not to be detected as fake. By studying the rewards values over time, we can see that the agent can generate news content which the adversary cannot easily detect as fake. The trained adversary has an accuracy of $\%81.3$ and AUC of $\%75.3$. 


In Table~\ref{table:examples} you can see several examples of generated news, given the topic. The provided examples in this table illustrate the performance of our method in generating synthetic news content. The first $10$ words in bold are the topic given to the model. The model then generates the rest of the news, which is related to the topic and have a similar gossipy style as the provided dataset. 

Furthermore, in Figure~\ref{fig:rewards-conf} we show the reverse confidence of classifier ($1 - C_f$) of the fake news classifier for several periods of training iterations. This figure shows the confidence for the final generated news at terminal time $T$. From this figure we conclude that the agent can generate realistic fake news. 

\subsection{RL Rewards Convergence}

To evaluate the convergence of rewards in reinforcement learning, we show the rewards over each episode during the training phase. Figure~\ref{fig:rewards-conf} shows the mean reward for each episode over for each iteration. The results indicates that the average reward of agent is increasing over time, meaning that the agent is learning a policy $\pi(s)$ which can result in larger reward values. At first the rewards are low which is as a result of randomness during early episodes, but it increases as the agent learns better actions for each  state $s$. Note that in Figure~\ref{fig:rewards-conf} we only show the reward values for first $4000$ iterations to show its increasing behaviour.
Furthermore, from this convergence plot we can conclude that our model can be trained faster than other baselines such as Grover or GPT-2. Our model tries to leverage an already trained language model to generate news content, which takes less than a day to train, while other methods try to learn a language model from scratch which is time consuming and can take up to weeks~\cite{zellers2019defending}. 

\subsection{Reward Function Parameter Analysis}
The reward function in our proposed method has three parameters $\alpha$, $\beta$, and $\lambda$ for changing the effect of each term in Equation~\ref{eq:reward_2}. We illustrate the effect of these parameters by changing them as $\alpha, \beta, \lambda \in \{0.0, 0.25, 0.5, 0.75, 1.0\}$ and calculating the ROUGE-L score. Figure~\ref{fig:params} shows the effect of each component on the ROUGE-L score. For each parameter, we consider other parameter values as $0.5$. While $\alpha$ and $\lambda$ have little effect on the ROUGE-L score, the $\beta$ parameter has a larger impact as it considers how well the generated news content overlaps with the given dataset $\mathcal{X}$.

\section{Conclusion and Future Work}
\label{section_conclution}

Text generation is a crucial task 
in different NLP applications. One application of text generation is news content generation. Current language models are very broad and they cannot generate news related to a topic. To this end, we proposed a reinforced model RLTG to control a language model toward news content generation. This model uses Deep Q-Learning to train an agent capable of selecting words from a language model's output in order to the generated news is related to a given topic. 

There are many future directions for this problem. One future direction is to train a model capable of generating only true news in order to write true article news. In addition, we can study the hidden features and differences between real and synthetic news content. This can greatly help us to detect machine generated fake news in future. Third, we will be able explore how to differentiate the human-written news and machine-generated news to help better detect fake news.

\begin{table*}[t!]
    \centering
    \begin{tabularx}{\linewidth}{c  c  X}
        \toprule
        {\bf \#} & {\bf Measure} & {\bf Question } \\ \midrule
        1 &  Style & \begin{minipage}{0.85\textwidth} Is the style of this article consistent? {\bf (3)}. Yes, this sounds like an article I would find at an online news source. {\bf (2)}. Sort of, but there are certain sentences that are awkward or strange. {\bf (1)}. No, it reads like it’s written by a madman.\end{minipage}\\ \hline \\ [-1.8ex]
        2 &  Content & \begin{minipage}{0.85\textwidth}Does the content of this article make sense? {\bf (3)}. Yes, this article reads coherently. {\bf (2)}. Sort of, but I don’t understand what the author means in certain places. {\bf (1)}. No, I have no (or almost no) idea what the author is trying to say.\end{minipage}\\ \hline \\ [-1.8ex]
        3 &  Title &  \begin{minipage}{0.85\textwidth}Does the article sound like it’s around a topic? {\bf (3)}. Yes, I feel that this article is talking about a single topic. {\bf (2)}. Sort of, I’m not sure what the article is about. {\bf (1)}. No, it seems this article is gibberish.\end{minipage}\\ \hline \\ [-1.8ex]
        4 &  Overall & \begin{minipage}{0.85\textwidth}Does the article read like it comes from a trustworthy source? {\bf (3)}. Yes, I feel that this article could come from a news source I would trust. {\bf (2)}. Sort of, but something seems a bit fishy. {\bf (1)}. No, this seems like it comes from an unreliable source.\end{minipage} \\ \bottomrule
    \end{tabularx}
    \caption{Human evaluation questionnaire}
    \label{tab:my_label}
\end{table*}
\bibliographystyle{ACM-Reference-Format}
\bibliography{main}


\appendix
\section{Human Evaluation}
\label{appx:A}

\autoref{tab:my_label} shows the designed questionnaire for evaluating the performance of the language models. We used a similar measure as~\cite{zellers2019defending}.

\end{document}